# Machine Transliteration

**Kevin Knight and Jonathan Graehl**
Information Sciences Institute
University of Southern California
Marina del Rey, CA 90292
knight@isi.edu, graehl@isi.edu

## Abstract

It is challenging to translate names and technical terms across languages with different alphabets and sound inventories. These items are commonly transliterated, i.e., replaced with approximate phonetic equivalents. For example, *computer* in English comes out as コンピューター (konpyuutaa) in Japanese. Translating such items from Japanese back to English is even more challenging, and of practical interest, as transliterated items make up the bulk of text phrases not found in bilingual dictionaries. We describe and evaluate a method for performing backwards transliterations by machine. This method uses a generative model, incorporating several distinct stages in the transliteration process.

## 1 Introduction

Translators must deal with many problems, and one of the most frequent is translating proper names and technical terms. For language pairs like Spanish/English, this presents no great challenge: a phrase like *Antonio Gil* usually gets translated as *Antonio Gil*. However, the situation is more complicated for language pairs that employ very different alphabets and sound systems, such as Japanese/English and Arabic/English. Phonetic translation across these pairs is called transliteration. We will look at Japanese/English transliteration in this paper.

Japanese frequently imports vocabulary from other languages, primarily (but not exclusively) from English. It has a special phonetic alphabet called *katakana*, which is used primarily (but not exclusively) to write down foreign names and loanwords. To write a word like *golfbag* in katakana, some compromises must be made. For example, Japanese has no distinct L and R sounds: the two English sounds collapse onto the same Japanese sound. A similar compromise must be struck for English H and F. Also, Japanese generally uses an alternating consonant-vowel structure, making it impossible to pronounce LFB without intervening vowels. Katakana writing is a syllabary rather than an alphabet—there is one symbol for ga (ガ), another for gi (ギ), another for gu (グ), etc. So the way to write *golfbag* in katakana is ゴルフバッグ, roughly pronounced goruhubaggu. Here are a few more examples:

*Angela Johnson*
アンジラ・ジョンソン
(anjira jyonson)

*New York Times*
ニューヨーク・タイムズ
(nyuuyooku taimuzu)

*ice cream*
アイスクリーム
(aisukuriimu)

Notice how the transliteration is more phonetic than orthographic; the letter *h* in *Johnson* does not produce any katakana. Also, a dot-separator (・) is used to separate words, but not consistently. And transliteration is clearly an information-losing operation: aisukuriimu loses the distinction between *ice cream* and *I scream*.

Transliteration is not trivial to automate, but we will be concerned with an even more challenging problem—going from katakana back to English, i.e., *back-transliteration*. Automating back-transliteration has great practical importance in Japanese/English machine translation. Katakana phrases are the largest source of text phrases that do not appear in bilingual dictionaries or training corpora (a.k.a. "not-found words"). However, very little computational work has been done in this area; (Yamron et al., 1994) briefly mentions a pattern-matching approach, while (Arbabi et al., 1994) discuss a hybrid neural-net/expert-system approach to (forward) transliteration.

The information-losing aspect of transliteration makes it hard to invert. Here are some problem instances, taken from actual newspaper articles:[1]

[1]Texts used in ARPA Machine Translation evaluations, November 1994.

?
アースデー
(aasudee)

?
ロバート・ショーン・レナード
(robaato shyoon renaado)

?
マスターズトーナメント
(masutaazutoonamento)

English translations appear later in this paper.

Here are a few observations about back-transliteration:

- Back-transliteration is less forgiving than transliteration. There are many ways to write an English word like *switch* in katakana, all equally valid, but we do not have this flexibility in the reverse direction. For example, we cannot drop the *t* in *switch*, nor can we write *arture* when we mean *archer*.
- Back-transliteration is harder than *romanization*, which is a (frequently invertible) transformation of a non-roman alphabet into roman letters. There are several romanization schemes for katakana writing—we have already been using one in our examples. Katakana writing follows Japanese sound patterns closely, so katakana often doubles as a Japanese pronunciation guide. However, as we shall see, there are many spelling variations that complicate the mapping between Japanese sounds and katakana writing.
- Finally, not all katakana phrases can be "sounded out" by back-transliteration. Some phrases are shorthand, e.g., ワープロ (waapuro) should be translated as *word processing*. Others are onomatopoetic and difficult to translate. These cases must be solved by techniques other than those described here.

The most desirable feature of an automatic back-transliterator is accuracy. If possible, our techniques should also be:

- portable to new language pairs like Arabic/English with minimal effort, possibly reusing resources.
- robust against errors introduced by optical character recognition.
- relevant to speech recognition situations in which the speaker has a heavy foreign accent.
- able to take textual (topical/syntactic) context into account, or at least be able to return a ranked list of possible English translations.

Like most problems in computational linguistics, this one requires full world knowledge for a 100% solution. Choosing between *Katarina* and *Catalina* (both good guesses for カタリナ) might even require detailed knowledge of geography and figure skating. At that level, human translators find the problem quite difficult as well, so we only aim to match or possibly exceed their performance.

## 2 A Modular Learning Approach

Bilingual glossaries contain many entries mapping katakana phrases onto English phrases, e.g.: (*aircraft carrier* $\leftrightarrow$ エアクラフトキャリア). It is possible to automatically analyze such pairs to gain enough knowledge to accurately map new katakana phrases that come along, and learning approach travels well to other languages pairs. However, a naive approach to finding direct correspondences between English letters and katakana symbols suffers from a number of problems. One can easily wind up with a system that proposes *iskrym* as a back-transliteration of aisukuriimu. Taking letter frequencies into account improves this to a more plausible-looking *isclim*. Moving to real words may give *is crime*: the *i* corresponds to ai, the *s* corresponds to su, etc. Unfortunately, the correct answer here is *ice cream*. After initial experiments along these lines, we decided to step back and build a generative model of the transliteration process, which goes like this:

1. An English phrase is written.
2. A translator pronounces it in English.
3. The pronunciation is modified to fit the Japanese sound inventory.
4. The sounds are converted into katakana.
5. Katakana is written.

This divides our problem into five sub-problems. Fortunately, there are techniques for coordinating solutions to such sub-problems, and for using generative models in the reverse direction. These techniques rely on probabilities and Bayes' Rule. Suppose we build an English phrase generator that produces word sequences according to some probability distribution $P(w)$. And suppose we build an English pronouncer that takes a word sequence and assigns it a set of pronunciations, again probabilistically, according to some $P(p|w)$. Given a pronunciation $p$, we may want to search for the word sequence $w$ that maximizes $P(w|p)$. Bayes' Rule lets us equivalently maximize $P(w) \cdot P(p|w)$, exactly the two distributions we have modeled.

Extending this notion, we settled down to build five probability distributions:

1. $P(w)$ — generates written English word sequences.
2. $P(e|w)$ — pronounces English word sequences.
3. $P(j|e)$ — converts English sounds into Japanese sounds.

4. P($k|j$) — converts Japanese sounds to katakana writing.
5. P($o|k$) — introduces misspellings caused by optical character recognition (OCR).

Given a katakana string $o$ observed by OCR, we want to find the English word sequence $w$ that maximizes the sum, over all $e$, $j$, and $k$, of

$$P(w) \cdot P(e|w) \cdot P(j|e) \cdot P(k|j) \cdot P(o|k)$$

Following (Pereira et al., 1994; Pereira and Riley, 1996), we implement P($w$) in a weighted finite-state acceptor (WFSA) and we implement the other distributions in weighted finite-state transducers (WFSTs). A WFSA is an state/transition diagram with weights and symbols on the transitions, making some output sequences more likely than others. A WFST is a WFSA with a pair of symbols on each transition, one input and one output. Inputs and outputs may include the empty symbol $\epsilon$. Also following (Pereira and Riley, 1996), we have implemented a general composition algorithm for constructing an integrated model P($x|z$) from models P($x|y$) and P($y|z$), treating WFSAs as WFSTs with identical inputs and outputs. We use this to combine an observed katakana string with each of the models in turn. The result is a large WFSA containing all possible English translations. We use Dijkstra's shortest-path algorithm (Dijkstra, 1959) to extract the most probable one.

The approach is modular. We can test each engine independently and be confident that their results are combined correctly. We do no pruning, so the final WFSA contains every solution, however unlikely. The only approximation is the Viterbi one, which searches for the best path through a WFSA instead of the best sequence (i.e., the same sequence does not receive bonus points for appearing more than once).

## 3 Probabilistic Models

This section describes how we designed and built each of our five models. For consistency, we continue to print written English word sequences in italics (*golf ball*), English sound sequences in all capitals (`G AA L F B AO L`), Japanese sound sequences in lower case (`g o r u h u b o o r u`) and katakana sequences naturally (ゴルフボール).

### 3.1 Word Sequences

The first model generates scored word sequences, the idea being that *ice cream* should score higher than *ice creme*, which should score higher than *aice kreem*. We adopted a simple unigram scoring method that multiplies the scores of the known words and phrases in a sequence. Our 262,000-entry frequency list draws its words and phrases from the Wall Street Journal corpus, an online English name list, and an online gazeteer of place names.[2] A portion of the WFSA looks like this:

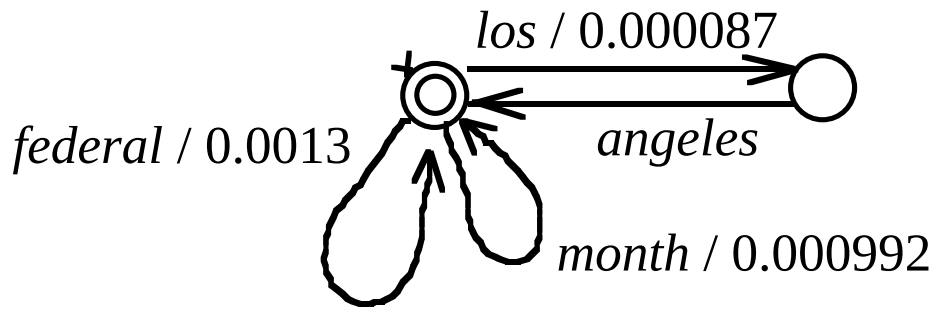

An ideal word sequence model would look a bit different. It would prefer exactly those strings which are actually grist for Japanese transliterators. For example, people rarely transliterate auxiliary verbs, but surnames are often transliterated. We have approximated such a model by removing high-frequency words like *has*, *an*, *are*, *am*, *were*, *their*, and *does*, plus unlikely words corresponding to Japanese sound bites, like *coup* and *oh*.

We also built a separate word sequence model containing only English first and last names. If we know (from context) that the transliterated phrase is a personal name, this model is more precise.

### 3.2 Words to English Sounds

The next WFST converts English word sequences into English sound sequences. We use the English phoneme inventory from the online CMU Pronunciation Dictionary,[3] minus the stress marks. This gives a total of 40 sounds, including 14 vowel sounds (e.g., `AA`, `AE`, `UW`), 25 consonant sounds (e.g., `K`, `HH`, `R`), plus our special symbol (`PAUSE`). The dictionary has pronunciations for 110,000 words, and we organized a phoneme-tree based WFST from it:

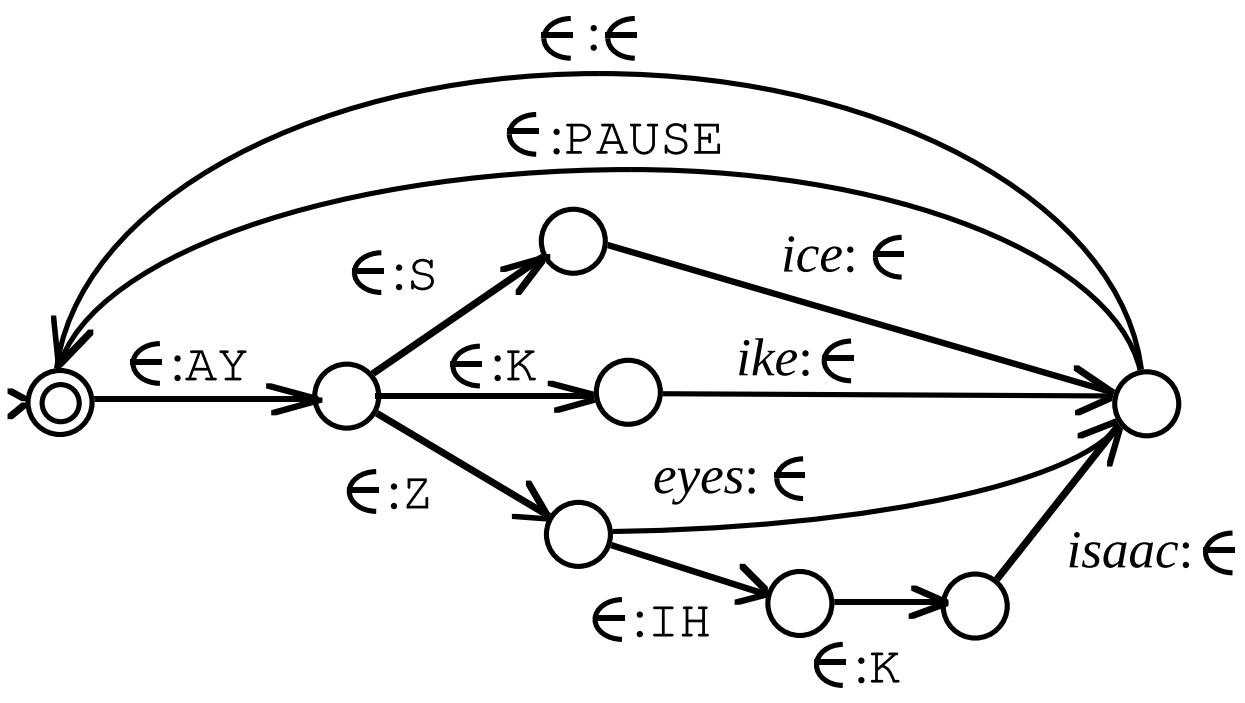

Note that we insert an optional `PAUSE` between word pronunciations. Due to memory limitations, we only used the 50,000 most frequent words.

We originally thought to build a general letter-to-sound WFST, on the theory that while wrong (overgeneralized) pronunciations might occasionally be generated, Japanese transliterators also mispronounce words. However, our letter-to-sound WFST did not match the performance of Japanese translit-erators, and it turns out that mispronunciations are modeled adequately in the next stage of the cascade.

[2] Available from the ACL Data Collection Initiative.

[3] `http://www.speech.cs.cmu.edu/cgi-bin/cmudict`.

### 3.3 English Sounds to Japanese Sounds

Next, we map English sound sequences onto Japanese sound sequences. This is an inherently information-losing process, as English `R` and `L` sounds collapse onto Japanese `r`, the 14 English vowel sounds collapse onto the 5 Japanese vowel sounds, etc. We face two immediate problems:

1. What is the target Japanese sound inventory?
2. How can we build a WFST to perform the sequence mapping?

An obvious target inventory is the Japanese syllabary itself, written down in katakana (e.g., ニ) or a roman equivalent (e.g., `ni`). With this approach, the English sound `K` corresponds to one of カ (`ka`), キ (`ki`), ク (`ku`), ケ (`ke`), or コ (`ko`), depending on its context. Unfortunately, because katakana is a syllabary, we would be unable to express an obvious and useful generalization, namely that English `K` usually corresponds to Japanese `k`, independent of context. Moreover, the correspondence of Japanese katakana writing to Japanese sound sequences is not perfectly one-to-one (see next section), so an independent sound inventory is well-motivated in any case. Our Japanese sound inventory includes 39 symbols: 5 vowel sounds, 33 consonant sounds (including doubled consonants like `kk`), and one special symbol (`pause`). An English sound sequence like (`P R OW PAUSE S AA K ER`) might map onto a Japanese sound sequence like (`p u r o pause s a kk a a`). Note that long Japanese vowel sounds are written with two symbols (`a a`) instead of just one (`aa`). This scheme is attractive because Japanese sequences are almost always longer than English sequences.

Our WFST is learned automatically from 8,000 pairs of English/Japanese sound sequences, e.g., `((S AA K ER)` ↔ `(s a kk a a))`. We were able to produce these pairs by manipulating a small English-katakana glossary. For each glossary entry, we converted English words into English sounds using the previous section's model, and we converted katakana words into Japanese sounds using the next section's model. We then applied the estimation-maximization (EM) algorithm (Baum, 1972) to generate symbol-mapping probabilities, shown in Figure 1. Our EM training goes like this:

1. For each English/Japanese sequence pair, compute all possible *alignments* between their elements. In our case, an alignment is a drawing that connects each English sound with one or more Japanese sounds, such that all Japanese sounds are covered and no lines cross. For example, there are two ways to align the pair `((L OW) <-> (r o o))`:

```
L   OW          L   OW
|   /\         /\   |
r  o  o       r  o  o
```

2. For each pair, assign an equal weight to each of its alignments, such that those weights sum to 1. In the case above, each alignment gets a weight of 0.5.
3. For each of the 40 English sounds, count up instances of its different mappings, as observed in all alignments of all pairs. Each alignment contributes counts in proportion to its own weight.
4. For each of the 40 English sounds, normalize the scores of the Japanese sequences it maps to, so that the scores sum to 1. These are the symbol-mapping probabilities shown in Figure 1.
5. Recompute the alignment scores. Each alignment is scored with the product of the scores of the symbol mappings it contains.
6. Normalize the alignment scores. Scores for each pair's alignments should sum to 1.
7. Repeat 3-6 until the symbol-mapping probabilities converge.

We then build a WFST directly from the symbol-mapping probabilities:

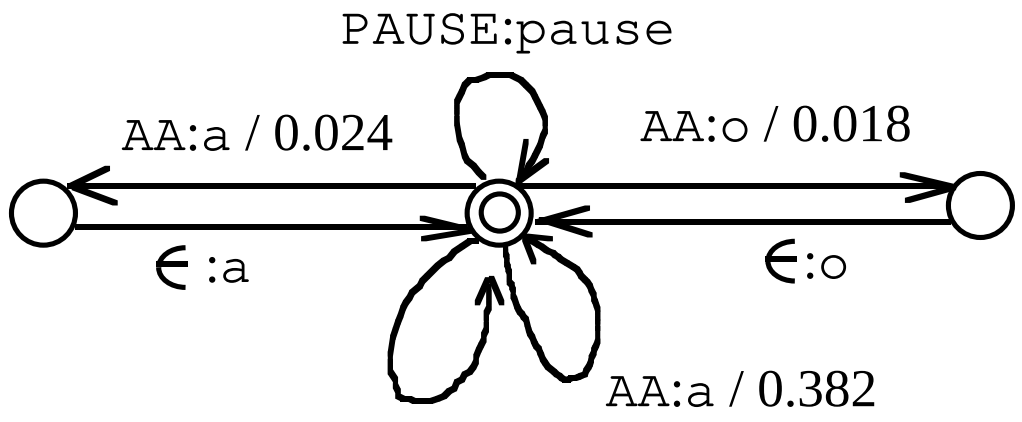

Our WFST has 99 states and 283 arcs.

We have also built models that allow individual English sounds to be "swallowed" (i.e., produce zero Japanese sounds). However, these models are expensive to compute (many more alignments) and lead to a vast number of hypotheses during WFST composition. Furthermore, in disallowing "swallowing," we were able to automatically remove hundreds of potentially harmful pairs from our training set, e.g., `((B AA R B ER SH AA P)` ↔ `(b a a b a a))`. Because no alignments are possible, such pairs are skipped by the learning algorithm; cases like these must be solved by dictionary lookup anyway. Only two pairs failed to align when we wished they had—both involved turning English `Y UW` into Japanese `u`, as in `((Y UW K AH L EY L IY)` ↔ `(u k u r e r e))`.

Note also that our model translates each English sound without regard to context. We have built also context-based models, using decision trees recoded as WFSTs. For example, at the end of a word, English `T` is likely to come out as (`t o`) rather than (`t`). However, context-based models proved unnecessary

| $e$ | $j$ | $P(j \mid e)$ |
|---|---|---|
| AA | o | 0.566 |
| | a | 0.382 |
| | a a | 0.024 |
| | o o | 0.018 |
| AE | a | 0.942 |
| | y a | 0.046 |
| AH | a | 0.486 |
| | o | 0.169 |
| | e | 0.134 |
| | i | 0.111 |
| | u | 0.076 |
| AO | o | 0.671 |
| | o o | 0.257 |
| | a | 0.047 |
| AW | a u | 0.830 |
| | a w | 0.095 |
| | o o | 0.027 |
| | a o | 0.020 |
| | a | 0.014 |
| AY | a i | 0.864 |
| | i | 0.073 |
| | a | 0.018 |
| | a i y | 0.018 |
| B | b | 0.802 |
| | b u | 0.185 |
| CH | ch y | 0.277 |
| | ch | 0.240 |
| | tch i | 0.199 |
| | ch i | 0.159 |
| | tch | 0.038 |
| | ch y u | 0.021 |
| | tch y | 0.020 |
| D | d | 0.535 |
| | d o | 0.329 |
| | dd o | 0.053 |
| | j | 0.032 |
| DH | z | 0.670 |
| | z u | 0.125 |
| | j | 0.125 |
| | a z | 0.080 |
| EH | e | 0.901 |
| | a | 0.069 |
| ER | a a | 0.719 |
| | a | 0.081 |
| | a r | 0.063 |
| | e r | 0.042 |
| | o r | 0.029 |
| EY | e e | 0.641 |
| | a | 0.122 |
| | e | 0.114 |
| | e i | 0.080 |
| | a i | 0.014 |
| F | h | 0.623 |
| | h u | 0.331 |
| | hh | 0.019 |
| | a h u | 0.010 |
| G | g | 0.598 |
| | g u | 0.304 |
| | gg u | 0.059 |
| | gg | 0.010 |
| HH | h | 0.959 |
| | w | 0.014 |
| IH | i | 0.908 |
| | e | 0.071 |
| IY | i i | 0.573 |
| | i | 0.317 |
| | e | 0.074 |
| | e e | 0.016 |
| JH | j | 0.329 |
| | j y | 0.328 |
| | j i | 0.129 |
| | jj i | 0.066 |
| | e j i | 0.057 |
| | z | 0.032 |
| | g | 0.018 |
| | jj | 0.012 |
| | e | 0.012 |
| K | k | 0.528 |
| | k u | 0.238 |
| | kk u | 0.150 |
| | kk | 0.043 |
| | k i | 0.015 |
| | k y | 0.012 |
| L | r | 0.621 |
| | r u | 0.362 |
| M | m | 0.653 |
| | m u | 0.207 |
| | n | 0.123 |
| | n m | 0.011 |
| N | n | 0.978 |
| NG | n g u | 0.743 |
| | n | 0.220 |
| | n g | 0.023 |
| OW | o | 0.516 |
| | o o | 0.456 |
| | o u | 0.011 |
| OY | o i | 0.828 |
| | o o i | 0.057 |
| | i | 0.029 |
| | o i y | 0.029 |
| | o | 0.027 |
| | o o y | 0.014 |
| | o o | 0.014 |
| P | p | 0.649 |
| | p u | 0.218 |
| | pp u | 0.085 |
| | pp | 0.045 |
| PAUSE | pause | 1.000 |
| R | r | 0.661 |
| | a | 0.170 |
| | o | 0.076 |
| | r u | 0.042 |
| | u r | 0.016 |
| | a r | 0.012 |
| S | s u | 0.539 |
| | s | 0.269 |
| | sh | 0.109 |
| | u | 0.028 |
| | ss | 0.014 |
| SH | sh y | 0.475 |
| | sh | 0.175 |
| | ssh y u | 0.166 |
| | ssh y | 0.088 |
| | sh i | 0.029 |
| | ssh | 0.027 |
| | sh y u | 0.015 |
| T | t | 0.463 |
| | t o | 0.305 |
| | tt o | 0.103 |
| | ch | 0.043 |
| | tt | 0.021 |
| | ts | 0.020 |
| | ts u | 0.011 |
| TH | s u | 0.418 |
| | s | 0.303 |
| | sh | 0.130 |
| | ch | 0.038 |
| | t | 0.029 |
| UH | u | 0.794 |
| | u u | 0.098 |
| | dd | 0.034 |
| | a | 0.030 |
| | o | 0.026 |
| UW | u u | 0.550 |
| | u | 0.302 |
| | y u u | 0.109 |
| | y u | 0.021 |
| V | b | 0.810 |
| | b u | 0.150 |
| | w | 0.015 |
| W | w | 0.693 |
| | u | 0.194 |
| | o | 0.039 |
| | i | 0.027 |
| | a | 0.015 |
| | e | 0.012 |
| Y | y | 0.652 |
| | i | 0.220 |
| | y u | 0.050 |
| | u | 0.048 |
| | b | 0.016 |
| Z | z | 0.296 |
| | z u | 0.283 |
| | j | 0.107 |
| | s u | 0.103 |
| | u | 0.073 |
| | a | 0.036 |
| | o | 0.018 |
| | s | 0.015 |
| | n | 0.013 |
| | i | 0.011 |
| | sh | 0.011 |
| ZH | j y | 0.324 |
| | sh i | 0.270 |
| | j i | 0.173 |
| | j | 0.135 |
| | a j y u | 0.027 |
| | sh y | 0.027 |
| | s | 0.027 |
| | a j i | 0.016 |

Figure 1: English sounds (in capitals) with probabilistic mappings to Japanese sound sequences (in lower case), as learned by estimation-maximization. Only mappings with conditional probabilities greater than 1% are shown, so the figures may not sum to 1.

for back-transliteration.[4] They are more useful for English-to-Japanese forward transliteration.

### 3.4 Japanese sounds to Katakana

To map Japanese sound sequences like (`m o o t a a`) onto katakana sequences like (モーター), we manually constructed two WFSTs. Composed together, they yield an integrated WFST with 53 states and 303 arcs. The first WFST simply merges long Japanese vowel sounds into new symbols `aa`, `ii`, `uu`, `ee`, and `oo`. The second WFST maps Japanese sounds onto katakana symbols. The basic idea is to consume a whole syllable worth of sounds before producing any katakana, e.g.:

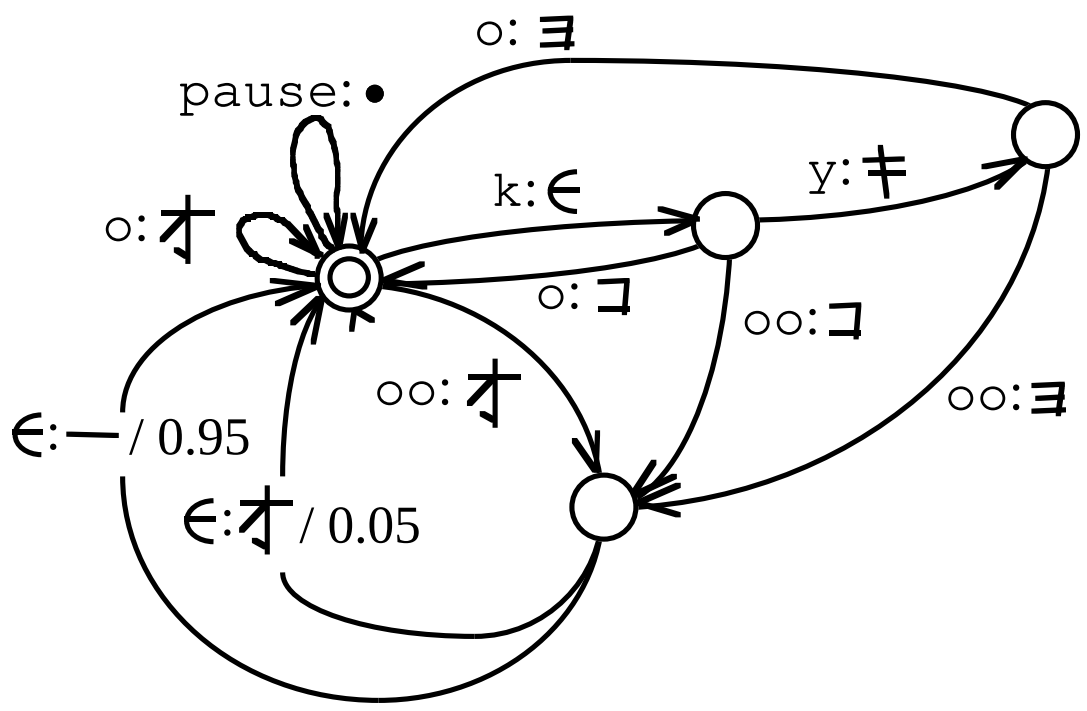

This fragment shows one kind of spelling variation in Japanese: long vowel sounds (`oo`) are usually written with a long vowel mark (オー) but are sometimes written with repeated katakana (オオ). We combined corpus analysis with guidelines from a Japanese textbook (Jorden and Chaplin, 1976) to turn up many spelling variations and unusual katakana symbols:

- the sound sequence (`j i`) is usually written ジ, but occasionally ヂ.
- (`g u a`) is usually グア, but occasionally グァ.
- (`w o o`) is variously ウオー, ウォー, or with a special, old-style katakana for `wo`.
- (`y e`) may be エ, イエ, or イェ.
- (`w i`) is either ウイ or ウィ.
- (`n y e`) is a rare sound sequence, but is written ニェ when it occurs.
- (`t y u`) is rarer than (`ch y u`), but is written テュ when it occurs.

and so on.

Spelling variation is clearest in cases where an English word like *switch* shows up transliterated variously (スィッチ, スイッチ, スウィッチ) in different dictionaries. Treating these variations as an equivalence class enables us to learn general sound mappings even if our bilingual glossary adheres to a single narrow spelling convention. We do not, however, generate all katakana sequences with this model; for example, we do not output strings that begin with a subscripted vowel katakana. So this model also serves to filter out some ill-formed katakana sequences, possibly proposed by optical character recognition.

### 3.5 Katakana to OCR

Perhaps uncharitably, we can view optical character recognition (OCR) as a device that garbles perfectly good katakana sequences. Typical confusions made by our commercial OCR system include ボ for ポ, チ for ナ, ア for ァ, and 7 for ブ. To generate pre-OCR text, we collected 19,500 characters worth of katakana words, stored them in a file, and printed them out. To generate post-OCR text, we OCR'd the printouts. We then ran the EM algorithm to determine symbol-mapping ("garbling") probabilities. Here is part of that table:

| $k$ | $o$ | $P(o \mid k)$ |
|---|---|---|
| ピ | ピ | 0.492 |
| | ビ | 0.434 |
| | ヒ | 0.042 |
| | 7 | 0.011 |
| ビ | ビ | 1.000 |
| ハ | ハ | 0.964 |
| | ノ、 | 0.036 |

This model outputs a superset of the 81 katakana symbols, including spurious quote marks, alphabetic symbols, and the numeral 7.

## 4 Example

We can now use the models to do a sample back-transliteration. We start with a katakana phrase as observed by OCR. We then serially compose it with the models, in reverse order. Each intermediate stage is a WFSA that encodes many possibilities. The final stage contains all back-transliterations suggested by the models, and we finally extract the best one.

We start with the `masutaazutoonamento` problem from Section 1. Our OCR observes:

マスクーズトーチメント

This string has two recognition errors: ク (`ku`) for タ (`ta`), and チ (`chi`) for ナ (`na`). We turn the string into a chained 12-state/11-arc WFSA and compose it with the $P(k|o)$ model. This yields a fatter 12-state/15-arc WFSA, which accepts the correct spelling at a lower probability. Next comes the $P(j|k)$ model, which produces a 28-state/31-arc WFSA whose highest-scoring sequence is:

`m a s u t a a z u t o o ch i m e n t o`

Next comes $P(e|j)$, yielding a 62-state/241-arc WFSA whose best sequence is:

`M AE S T AE AE DH UH T AO AO CH IH M EH N T AO`

[4] And harmfully restrictive in their unsmoothed incarnations.

Next to last comes $P(w|e)$, which results in a 2982-state/4601-arc WFSA whose best sequence (out of myriads) is:

*masters tone am ent awe*

This English string is closest phonetically to the Japanese, but we are willing to trade phonetic proximity for more sensical English; we rescore this WFSA by composing it with $P(w)$ and extract the best translation:

*masters tournament*

(Other Section 1 examples are translated correctly as *earth day* and *robert sean leonard.*)

## 5 Experiments

We have performed two large-scale experiments, one using a full-language $P(w)$ model, and one using a personal name language model.

In the first experiment, we extracted 1449 unique katakana phrases from a corpus of 100 short news articles. Of these, 222 were missing from an on-line 100,000-entry bilingual dictionary. We back-transliterated these 222 phrases. Many of the translations are perfect: *technical program*, *sex scandal*, *omaha beach*, *new york times*, *ramon diaz.* Others are close: *tanya harding*, *nickel simpson*, *danger washington*, *world cap.* Some miss the mark: *nancy care again*, *plus occur*, *patriot miss real.* While it is difficult to judge overall accuracy—some of the phases are onomatopoetic, and others are simply too hard even for good human translators—it is easier to identify system weaknesses, and most of these lie in the $P(w)$ model. For example, *nancy kerrigan* should be preferred over *nancy care again.*

In a second experiment, we took katakana versions of the names of 100 U.S. politicians, e.g.: ジョン・ブロー (`jyon.buroo`), アルホンス・ダマット (`arhonsu.damatto`), and マイク・デワイン (`maiku.dewain`). We back-transliterated these by machine and asked four human subjects to do the same. These subjects were native English speakers and news-aware; we gave them brief instructions, examples, and hints. The results were as follows:

| | human | machine |
|---|---|---|
| correct (e.g., *spencer abraham* / *spencer abraham*) | 27% | 64% |
| phonetically equivalent, but misspelled (e.g., *richard brian* / *richard bryan*) | 7% | 12% |
| incorrect (e.g., *olin hatch* / *orren hatch*) | 66% | 24% |

There is room for improvement on both sides. Being English speakers, the human subjects were good at English name spelling and U.S. politics, but not at Japanese phonetics. A native Japanese speaker might be expert at the latter but not the former. People who are expert in all of these areas, however, are rare.

On the automatic side, many errors can be corrected. A first-name/last-name model would rank *richard bryan* more highly than *richard brian.* A bigram model would prefer *orren hatch* over *olin hatch.* Other errors are due to unigram training problems, or more rarely, incorrect or brittle phonetic models. For example, "Long" occurs much more often than "Ron" in newspaper text, and our word selection does not exclude phrases like "Long Island." So we get *long wyden* instead of *ron wyden.* Rare errors are due to incorrect or brittle phonetic models.

Still the machine's performance is impressive. When word separators ( • ) are removed from the katakana phrases, rendering the task exceedingly difficult for people, the machine's performance is unchanged. When we use OCR, 7% of katakana tokens are mis-recognized, affecting 50% of test strings, but accuracy only drops from 64% to 52%.

## 6 Discussion

We have presented a method for automatic back-transliteration which, while far from perfect, is highly competitive. It also achieves the objectives outlined in Section 1. It ports easily to new language pairs; the $P(w)$ and $P(e|w)$ models are entirely reusable, while other models are learned automatically. It is robust against OCR noise, in a rare example of high-level language processing being useful (necessary, even) in improving low-level OCR.

We plan to replace our shortest-path extraction algorithm with one of the recently developed $k$-shortest path algorithms (Eppstein, 1994). We will then return a ranked list of the $k$ best translations for subsequent contextual disambiguation, either by machine or as part of an interactive man-machine system. We also plan to explore probabilistic models for Arabic/English transliteration. Simply identifying which Arabic words to transliterate is a difficult task in itself; and while Japanese tends to insert extra vowel sounds, Arabic is usually written without any (short) vowels. Finally, it should also be possible to embed our phonetic shift model $P(j|e)$ inside a speech recognizer, to help adjust for a heavy Japanese accent, although we have not experimented in this area.

## 7 Acknowledgments

We would like to thank Alton Earl Ingram, Yolanda Gil, Bonnie Glover-Stalls, Richard Whitney, and Kenji Yamada for their helpful comments. We would also like to thank our sponsors at the Department of Defense.